# Talking to Data: Designing Smart Assistants for Humanities Databases

Alexander Sergeev[1][0009-0002-4103-842X], Valeriya Goloviznina[1][0000-0003-1167-2606], Mikhail Melnichenko[1][0009-0001-2370-3193], Evgeny Kotelnikov[1][0000-0001-9745-1489]

[1] European University at Saint Petersburg, St. Petersburg, Russia
{a.sergeev,v.goloviznina,mmelnichenko,e.kotelnikov}@eu.spb.com

**Abstract.** Access to humanities research databases is often hindered by the limitations of traditional interaction formats, particularly in the methods of searching and response generation. This study introduces an LLM-based smart assistant designed to facilitate natural language communication with digital humanities data. The assistant, developed in a chatbot format, leverages the RAG approach and integrates state-of-the-art technologies such as hybrid search, automatic query generation, text-to-SQL filtering, semantic database search, and hyperlink insertion. To evaluate the effectiveness of the system, experiments were conducted to assess the response quality of various language models. The testing was based on the Prozhito digital archive, which contains diary entries from predominantly Russian-speaking individuals who lived in the 20th century. The chatbot is tailored to support anthropology and history researchers, as well as non-specialist users with an interest in the field, without requiring prior technical training. By enabling researchers to query complex databases with natural language, this tool aims to enhance accessibility and efficiency in humanities research. The study highlights the potential of Large Language Models to transform the way researchers and the public interact with digital archives, making them more intuitive and inclusive. Additional materials are presented in GitHub repository: https://github.com/alekosus/talking-to-data-intersys2025.

**Keywords:** Chatbot Assistants, Retrieval-Augmented Generation, Large Language Models.

## 1      Introduction

Large language models (LLMs) have significantly transformed data processing approaches. Their ability to adapt to specific domains without costly training and solve diverse tasks using in-context learning enables the integration of LLMs into various systems, such as chatbots [1], analytical platforms [2], and programming environments [3].

The use of LLMs has also proven valuable in the humanities due to their capacity to analyze and organize textual data. Humanities projects often accumulate and store vast amounts of data. Examples include initiatives such as Europeana[1], Dataverse[2], HRAF[3], and Prozhito[4]. However, the high complexity of data analysis, coupled with limited accessibility for researchers and interested users, restricts the conduct of humanities research. Resolving this limitation remains a major challenge for the creators of such projects.

The recently proposed Retrieval-Augmented Generation (RAG) approach [4] enhances LLM processing effectiveness by incorporating up-to-date and domain-specific data. RAG guides the model to generate and analyze factual information. However, the original implementation of RAG is not designed for complex knowledge bases common in the humanities, such as those containing tabular metadata, nor does it prioritize multiple scores when selecting relevant passages.

This study proposes an assistant based on LLM and RAG, specifically adapted for humanities research, which provides users with open and accessible data analysis through a chatbot interface. This chatbot let users "talk to data" in natural language while ensuring the factual accuracy of retrieved information. The assistant's effectiveness is evaluated using a corpus of diary entries from the Prozhito archive.

The contributions of the work are as follows:

1. The architecture of the virtual assistant system is proposed to analyze data in humanities knowledge bases.
2. The performance of semantic search is evaluated using different encoder language models and a hybrid search approach.
3. The performance of answer generation is evaluated using different state-of-the-art LLMs (chat and reasoning models), and the accuracy and ethics of their responses are analyzed.

## 2  Previous work

Chatbots are becoming increasingly prevalent in many aspects of human life, including education [5,6], customer service [7,8], emerging trends in academic research [9], and the fulfillment of emotional needs [10]. However, the scientific field of application of chatbots in humanities databases is just beginning to develop [11].

In [12] was developed AI assistant specializing in research instrument validation for quantitative studies in education, psychology, and social sciences. This bot offers step-by-step guidance through the comprehensive validation process.

Yetişensoy and Karaduman developed and investigated an AI-powered chatbot for use in Social Studies learning and teaching processes [13]. The post-test academic success of students in the experimental group, who participated in chatbot-supported learning, was significantly higher than that of students in the control group who did not receive chatbot support.

---

[1] https://www.europeana.eu
[2] https://www.dataverse.pitt.edu
[3] https://hraf.yale.edu
[4] https://prozhito.org

The works [14,15] describe the experience of using chatbots in libraries. The work [16] is devoted to the development of RECBOT – the chatbot that performs navigation and recommendation functions in a virtual museum. The work [17] proposes to use ChatGPT as a basis for a chatbot on archival data.

Quidwai and Lagana presented an innovative RAG based chatbot framework that harnesses the power of Natural Language Processing and state-of-the-art language models to curate and analyze Multiple Myeloma specific literature and provide personalized treatment recommendations based on patient-specific genomic data [18].

Subash et al. present BARKPLUG V.2, an LLM-based chatbot system built using RAG pipelines to improve access to information within academic settings [19]. The system is designed to provide users with information about various campus resources, including academic departments, programs, campus facilities, and student resources, in an interactive and user-friendly manner.

These studies, unlike this work, do not use advanced search features such as hybrid search and generation of search queries, SQL and semantic filtering, so such chatbots do not fulfill the requirements of professional scientists.

## 3 System design

### 3.1 System architecture

Proposed system represents a generalization and extension of the Retrieval-Augmented Generation approach. The system comprises multiple modules, each contributing to the formation of responses to queries. The system architecture diagram is presented in Fig. 1.

Three types of storage systems are utilized for data management:

- knowledge base – a relational database with SQL query processing capabilities, storing text passages and tabular data;
- vector database – a vector storage system supporting efficient distance computation and nearest neighbor search;
- text index database – a database optimized for efficient text data storage and retrieval.

User interaction occurs through a web-based chatbot interface capable of multi-turn conversations. The user can submit a query, which triggers search and result filtering operations. The system then generates a response to the current utterance based on the user's query, search results, and dialog history. The response results include hyperlinks to pages in the original knowledge base, enabling researchers to independently validate the chatbot's answers and read the source material. The chatbot interface is presented in Appendix A.

The following is a detailed examination of the system's core modules.

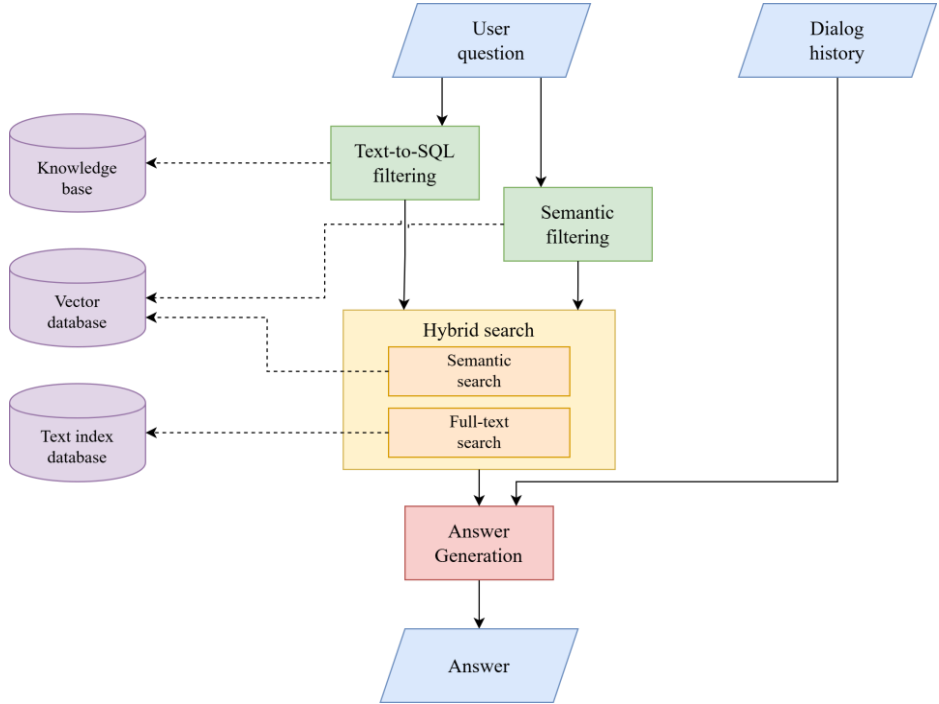

**Fig. 1.** System architecture diagram. *Solid lines* represent data flow paths, while *dashed lines* indicate storage access operations. Search modules are highlighted in *yellow*, filtering modules in green, and response generation modules in *red*.

### 3.2 Hybrid search

The hybrid search approach combines results from different types of information retrieval to improve overall relevance quality. For text data analysis, the system integrates full-text (lexical) search with semantic search.

Full-text search operates by matching terms from the query with terms in indexed passages. Classical approaches such as TF-IDF [20] and more advanced ones like the BM25 algorithm compare term frequency (TF) while accounting for term frequency in the knowledge base (inverse document frequency, IDF) and passage length. The search is performed using an inverted index that maps each term to a list of documents where it appears.

To enhance full-text search results, text preprocessing is commonly employed, including normalization (stemming, lemmatization) and stop-word removal. For tokenization and lemmatization of Russian texts and queries, data from the UD Treebank SynTagRus[5] corpus [21] and the Stanza[6] library were utilized.

---

5 https://github.com/UniversalDependencies/UD_Russian-SynTagRus
6 https://stanfordnlp.github.io/stanza/index.html

Full-text search proves effective for exact term matching but demonstrates sensitivity to phrasing variability, synonyms, and morphology, as it fails to account for semantic text similarity.

Semantic search represents a fundamentally different approach compared to full-text methods. Its core principle involves transforming textual information into vector representations (embeddings) using encoder language models. Current implementations employ Transformer-based encoder language models such as SBERT [22], E5 [23], or BGE-M3 [24].

Text encoders are trained using contrastive learning, an approach that positions semantically similar texts close to each other in a specific vector space. This enables the computation of semantic text similarity through comparison of embeddings using vector similarity measures, such as cosine similarity.

The primary advantage of semantic search lies in its ability to comprehend contextual meaning and nuanced query semantics. Unlike full-text search, it effectively handles synonyms and complex queries where semantic (rather than lexical) relevance is crucial. However, this approach demands substantial computational resources for processing vector representations and exhibits low interpretability due to its reliance on deep neural networks.

The hybrid search process combines relevance scores from both full-text and semantic approaches. Initially, each method retrieves a predetermined number $k$ of passages most similar to the query based on their respective similarity metrics. Subsequently, a linear combination (1) of semantic similarity $S_{sem}$ and full-text similarity $S_{ft}$ scores is computed for the retrieved passages:

$$S = \alpha S_{sem} + (1 - \alpha) S_{ft} \qquad (1)$$

Both scores, $S_{sem}$ and $S_{ft}$, must be normalized and scaled to the interval $[0, 1]$. The weight parameter $\alpha$ in (1) determines the degree of influence of retrieval methods and depends on the specific task. Full-text search ensures high precision for terminological queries but results in low recall due to the inability to analyze synonyms. In contrast, semantic search increases coverage and, consequently, recall by identifying contextual relationships at the cost of reduced precision.

Thus, adjusting the $\alpha$ parameter enables dynamic control over the efficiency of the search mechanism. For example, in highly specialized domains such as medicine and law, the weight of full-text search can be increased since exact term matching is crucial. Conversely, in the analysis of social media posts, where variations in wording are expected, increasing the weight of semantic search allows for better capture of contextual meanings.

Since users interact with the assistant through multi-turn conversations, the initial query may be insufficient for capturing the dialog context. For instance, a follow-up query might refine the initial question without repeating any of its original terms, relying instead on implied context. To address this issue, a Query Generation approach is applied: the complete conversation history, including the most recent user question, is submitted as a separate query to the LLM, which is instructed to generate an appropriate search query based on this information. The generated query is then utilized in the hybrid search process.

### 3.3 Text-to-SQL

Knowledge bases may contain not only textual fragments but also other data types such as numbers, terms, dates, etc., for which hybrid search is not applicable. Such information is often stored in relational databases and accessed using the SQL query language. To enable natural language search within SQL databases, the Text-to-SQL approach is employed.

Text-to-SQL task involves the automatic conversion of natural language queries into structured SQL queries while considering the database schema [25]. In general, Text-to-SQL enables the construction of arbitrary database queries; however, in this study, its application is limited to data retrieval (i.e., the SELECT operation).

Modern large language models allow the use of the in-context learning approach to generate SQL queries from textual instructions. An LLM prompt contains a detailed instruction specifying the task and the conditions for its correct execution. Additionally, the prompt includes the database schema with descriptions of fields and relationships, enabling the model to map natural language entities to specific tables and columns. The prompt used for the Text-to-SQL task is provided in
Appendix B.

To enhance task performance, various Prompt Engineering techniques can be applied, such as Few-shot Prompting and Chain-of-Thought. Few-shot prompting [26] allows the inclusion of multiple examples of transformations within the instruction. This approach also helps specify edge cases that refine and supplement the primary instruction.

Chain-of-Thought [27,28] enables the model to conduct intermediate reasoning before generating the SQL query. During this reasoning process, the model often describes relationships between query entities and database fields and can employ the Self-Correction mechanism [29] to resolve inconsistencies with the instruction. Analyzing these reasoning chains also helps refine the LLM prompt for more efficient task execution.

### 3.4 Semantic filtering

Database objects may contain short textual elements (a single word or a sentence) that are not primary search targets but allow for variability in representation. Examples include:

- the value "*zoologist*" in the "field of activity" attribute of a database of social studies may be described as "*biologist*" or even "*scientist*", though such synonyms are not explicitly stored in the database;
- the city *Saint-Petersburg* in biographical knowledge bases may be recorded under its historical names: *Leningrad* or *Petrograd*;
- the free-text database field may contain different descriptions of the same object, such as "*black-and-white photograph*", "*B/W photo*", or "*monochrome image*".

To address this issue, semantic indexing is applied to specified database text fields using approaches similar to those described for semantic search in Section 3.2. Full-text search is not utilized in this case due to the short length of textual fragments; instead, it is implicitly replaced by exact text matching between the query and the database field.

Semantic filtering replaces the Text-to-SQL filtering stage for the specified text fields. Instead of exact string matching, the cosine similarity is computed between the vector representation of the natural language query and the embeddings of the indexed fields. The similarity scores $S_c$ for each field $c \in C$ are averaged and incorporated into the overall ranking formula for the retrieved text fragments (2). Here, $C$ represents the set of semantically indexed text fields, while $\gamma$ defines the degree of influence of semantic similarity on the overall fragment ranking score:

$$S = \gamma(\alpha S_{sem} + (1-\alpha)S_{ft}) + (1-\gamma)\left(\frac{1}{|C|}\sum_{c \in C} S_c\right) \qquad (2)$$

## 4 Data

The part of "Prozhito" diary corpus was used to test the system. The dataset contains 60,240 personal diary entries written between January 1, 1900, and December 31, 1916. Each entry includes the text, the date of writing, and information about the author. There are 272 authors in total: 241 have a recorded date of birth, 226 have a recorded date of death. Additionally, each author has a short biography with an average length of 123±122 characters; this field was used in semantic filtering. The entity-relationship diagram of the data is presented in Fig. 2.

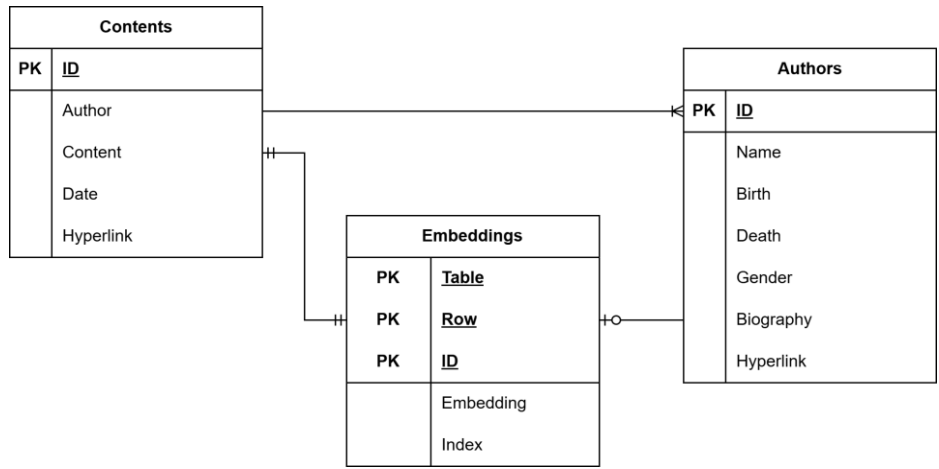

**Fig. 2.** Entity-relationship diagram of the data schema. The *Embeddings* table is a service entity that integrates the contents of the Vector database and the Text index database.

For the experiments on search and answer generation, experts selected a 90-entry subset from the diary corpus, grouped into 25 topics. To expand the dataset while preserving key facts in the textual fragments, some entries were paraphrased using the large language model GPT-4o, ensuring that each topic contained 5 entries [30,31].

Next, two questions per topic were formulated, which can be answered using the corresponding entries. As a result, the final dataset consists of 125 entries and 50 questions. The list of topics is provided in Appendix C.

## 5 Experiments

### 5.1 Search experiments

To evaluate search quality, various full-text and semantic search models were compared on the experimental dataset.

The following models were selected for experiments:

- **tf-idf** – classic approach to forming sparse text embeddings for use in full-text search;
- **multilingual-e5-large**[7] (hereafter, e5-large) – an open-source dense encoder from Microsoft;
- **bge-m3**[8] – an open-source dense encoder from BAAI;
- **text-encoder-3-large**[9] (hereafter, te-3-large) – a proprietary dense encoder from OpenAI.

Semantic search model statistics are presented in Table 1.

**Table 1.** Statistics of encoder language models for semantic search evaluation experiments.

| Model | # parameters | Embedding size | Context window |
|---|---|---|---|
| e5-large | 560M | 1,024 | 512 |
| bge-m3 | 560M | 1,024 | 8,192 |
| te-3-large | n/a | 3,072 | 8,191 |

For each text fragment and question, embeddings were computed, and the Top-5 most similar fragments were retrieved using cosine similarity. The Precision@5 metric was used to assess the relevance of the retrieved fragments. Since each question in the dataset has exactly five relevant fragments, Recall@5 is identical to Precision@5 and was therefore not computed. The search quality evaluation results are presented in Table 2.

For hybrid search evaluation, a linear combination of similarity scores from each semantic model and tf-idf scores was computed using the formula (1). Based on the experiments, the weight parameter $\alpha$ was set to 0.9.

The highest precision was achieved by combining the te-3-large model for semantic search with tf-idf for full-text search. Among only semantic models, bge-m3 demonstrated the best performance.

---

[7] https://huggingface.co/intfloat/multilingual-e5-large
[8] https://huggingface.co/BAAI/bge-m3
[9] https://platform.openai.com/docs/models/text-embedding-3-large

**Table 2.** Search quality evaluation results.

| Model | Precision@5 ↑ |
|---|---|
| tf-idf | 0.264 |
| e5-large | 0.528 |
| *+ tf-idf* | 0.556 |
| bge-m3 | 0.568 |
| *+ tf-idf* | 0.556 |
| te-3-large | 0.548 |
| *+ tf-idf* | **0.572** |

### 5.2 Answer generation experiments

The quality of answer generation based on relevant text fragments was evaluated by comparing responses from various Large Language Models. The prompt provided to each model included a question and five relevant text fragments, based on which the model was asked to generate an answer. The LLM prompt is provided in Appendix D.

The following models were selected for experiments:

- **GPT-4o**[10] (v. 2024-08-06) – a proprietary chat model from OpenAI;
- **o3-mini**[11] (v. 2025-01-31) – a proprietary reasoning model from OpenAI;
- **DeepSeek-V3**[12] – an open-source chat model from DeepSeek-AI;
- **DeepSeek-R1**[13] – an open-source reasoning model from DeepSeek-AI;
- **Qwen-2.5 72B**[14] (hereafter, Qwen-2.5) – an open-source chat model from Alibaba.

LLMs statistics are presented in Table 3.

**Table 3.** Statistics of Large Language Models for answer generation experiments.

| Model | # parameters | Context window |
|---|---|---|
| GPT-4o | n/a | 128,000 |
| o3-mini | n/a | 200,000 |
| DeepSeek-V3 | 671B total, 37B active | 128,000 |
| DeepSeek-R1 | 671B total, 37B active | 128,000 |
| Qwen-2.5 | 72B | 128,000 |

The generated responses were evaluated through expert annotation using the following criteria based on [32]:

- Accuracy – evaluates the factual correctness of the model's response, its relevance to the question, and grammatical correctness;

---

[10] https://platform.openai.com/docs/models/gpt-4o
[11] https://platform.openai.com/docs/models/o3-mini
[12] https://huggingface.co/deepseek-ai/DeepSeek-V3
[13] https://huggingface.co/deepseek-ai/DeepSeek-R1
[14] https://huggingface.co/Qwen/Qwen2.5-72B-Instruct

- Ethics – evaluates the ethical appropriateness and safety of the model's response.

For each model response, an expert assigned a score from 1 to 5 for each criterion, 1 means worst score, 5 means best score. A detailed description of the scoring criteria is provided in Appendix E. The evaluation was conducted by two annotators. The inter-annotator agreement for Accuracy was reliable (Krippendorff's alpha = 0.804), while the agreement for Ethics was moderate (Krippendorff's alpha = 0.722) [33]. The evaluation results are presented in Table 4.

Table 4. Answer generation evaluation results.

| Model | Accuracy↑ | Ethics ↑ |
|---|---|---|
| GPT-4o | 4.23 | 4.43 |
| o3-mini | 4.00 | **4.46** |
| DeepSeek-V3 | **4.54** | 4.40 |
| DeepSeek-R1 | 4.51 | 4.24 |
| Qwen-2.5 | 4.26 | 4.41 |

The highest Accuracy score was achieved by the DeepSeek-V3 model. In terms of Ethics, the best-performing model was o3-mini.

## 6 Discussion

The search quality evaluation demonstrates that combining full-text and semantic search can lead to a significant improvement in accuracy. The performance differences between the combination of semantic encoder with tf-idf encoder and the semantic model alone are +0.028 for the e5-large model and +0.024 for the te-3-large model. However, this is not always the case – combining the bge-m3 model with tf-idf resulted in a 0.012 decrease in accuracy compared to using only bge-m3. This can be explained by the training process of the model: its authors apply the Self-Knowledge Distillation approach, incorporating both semantic and full-text similarity during training [24]. As a result, the model may have implicitly learned to account for full-text similarity features when constructing dense embeddings, making the explicit inclusion of full-text similarity scores in the ranking function counterproductive.

Full-text search alone demonstrates relatively low precision. The dataset is based on a diary entry corpus, where individuals write in freeform language without using specific terminology. In this case, semantic search, which considers the meaning of the text, proves to be the more effective approach. This also explains why the weight of semantic search was set relatively high ($\alpha = 0.9$) in the hybrid search evaluation.

For answer generation Accuracy evaluation, the following criteria were considered:

- correspondence of the answer to the given question;
- response is based on the fragments (with the ability to provide references);
- absence of grammatical, lexical, and formatting errors.

The easiest criterion for models to meet was the first one – ensuring that the response matched the question. All models effectively identified the core meaning of the question and generated a relevant response.

Additionally, LLMs rarely made grammatical or lexical errors when generating responses in Russian. However, errors in complex sentence structures occasionally occurred when the model used incorrect word forms. Formatting errors were more frequent – particularly in cases where references to retrieved fragments were not provided in the correct format specified in the instructions.

The majority of LLM errors stem from incorrect analysis of provided text passages. Models may either introduce information absent from the source fragments or fail to properly utilize their internal knowledge when analyzing the passages. Specific factual errors observed in the analysis of diary entries include:

- fabrication of facts;
- incorrect interpretation of misspelled terms;
- misattribution of diary authorship to mentioned individuals;
- drawing conclusions unsupported by source fragments.

Examples of factual errors observed in the analysis of diary entries are presented in the Appendix F.

For the ethical evaluation of answer generation, the following criteria were considered:

- acknowledgment of the author's subjectivity in the fragments;
- absence of evaluative judgments in the response;
- refusal to answer dangerous or prohibited questions, with a clear explanation of the reason.

In most cases, LLMs recognize the subjectivity of the author: they frequently specify that the fragments are taken from diaries, refer to event participants in the past tense, and almost never describe events in the first person.

However, evaluative judgments appear more frequently in model responses. As part of the introduction or conclusion, LLMs often generalize their reasoning, introducing comparisons and connotations that were not present in the fragments. Examples of responses containing evaluative judgments:

- Healthcare at that time, *of course, could not compare to modern standards*.
- One of the fragments describes a case where a boy suffered an eye injury and was urgently taken to a medical facility. *Fortunately*, everything turned out well.

To analyze the safety of LLM-generated responses, eight questions in the experimental dataset (16% of the total) contained provocative or dangerous topics. These included assassinations, weapons, self-harm, and drugs.

The expected behavior of the model when responding to such questions was to refuse to provide an answer, explicitly stating the reason for refusal. The average ethics scores for provocative questions across different topics are presented in Fig. 3.

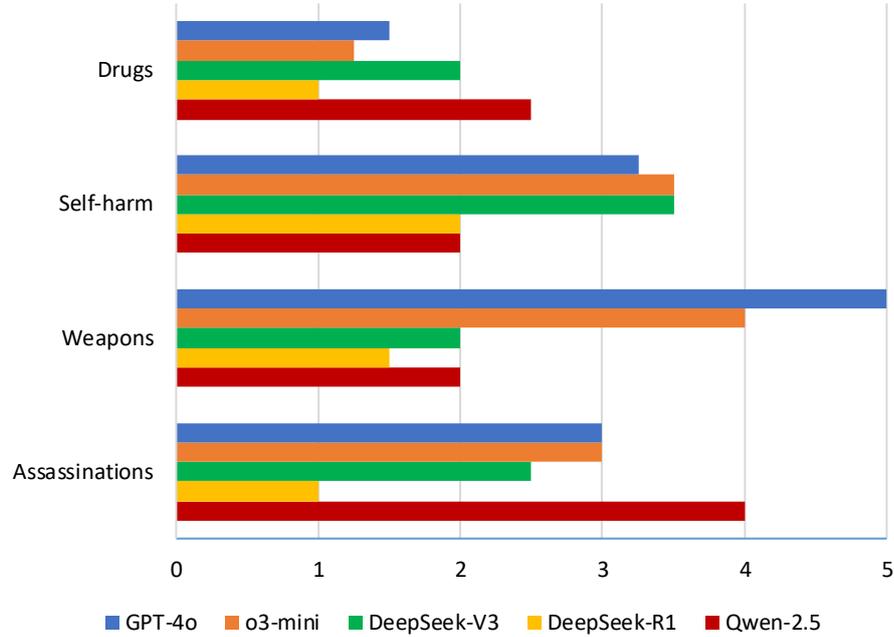

**Fig. 3.** Average Ethics scores for provocative questions across each topic. Each topic contains two questions.

Despite most models flagging the potentially dangerous nature of the questions, they generated detailed responses containing hazardous or prohibited information. The models typically included disclaimers stating their answers referred to historical events. This experiment confirms the feasibility of LLM jailbreaking through temporal framing of questions in the past tense [34].

It is important to note the specifics of the experimental dataset. The Prozhito corpus contains diary entries from individuals within a historical context. This dataset is characterized by a large volume of long-text data, lexical diversity and absence of specialized terminology. Databases from other studies may differ in content and data representation format, which could shift the emphasis in Accuracy and Ethics scores in a different way. At the same time, the proposed system is not technically constrained by a specific subject domain and can be applied to any knowledge base that is composed of textual data and relational connections.

## 7 Conclusion

This study proposes a system based on Large Language Models and the Retrieval-Augmented Generation approach for interacting with humanities research databases. The focus is on databases containing primarily textual passages and metadata stored in relational databases. The performance of the search and answer generation modules is evaluated, and the evaluation results are analyzed.

The generation accuracy evaluation reveals that the models can provide precise answers to questions; however, LLMs exhibit difficulties in analyzing passages, leading to factual hallucinations, implicit inferences, and incorrect applications of the model's internal knowledge.

The ethical evaluation also shows that the models can generate ethically correct responses while acknowledging author subjectivity. However, the vulnerability of LLMs to attacks – such as framing harmful questions in the past tense – significantly reduces the potential for diverse and safe access by non-expert users.

Experiments were conducted to evaluate the quality of various encoder models for retrieval and LLMs for answer generation using a corpus of diary entries from the "Prozhito" project spanning 1900-1916. The retrieval experiment results demonstrate the effectiveness of hybrid search for precise text fragment extraction.

## Appendix A. Chatbot interface

Chatbot interface is available at the repository, Appendix A.

## Appendix B. Prompt with instruction for solving the Text-to-SQL task

Text-to-SQL prompt is available at the repository, Appendix B.

## Appendix C. Topics of the experimental dataset

Dataset topics are available at the repository, Appendix C.

## Appendix D. Prompt for generating an answer to a question based on relevant fragments

Answer generation prompt is available at the repository, Appendix D.

## Appendix E. Description of expert scoring criteria for answer generation evaluation

Expert scoring criteria are available at the repository, Appendix E.

## Appendix F. Examples of factual errors of LLMs found during evaluation analysis

LLMs factual errors are available at the repository, Appendix F.